\documentclass[runningheads]{llncs} 
\usepackage[T1]{fontenc}
\usepackage{graphicx} 
\usepackage{amsmath,amsfonts}
\usepackage{algorithmic}
\usepackage{algorithm}
\usepackage{array}
\usepackage[caption=false,font=normalsize,labelfont=sf,textfont=sf]{subfig}
\usepackage{textcomp}
\usepackage{stfloats}
\usepackage{url}
\usepackage{verbatim}
\usepackage{graphicx}
\usepackage{cite}
\usepackage{multirow}
\usepackage{multicol}
\usepackage{booktabs}
\usepackage{wrapfig}
\usepackage{siunitx}
\usepackage{listings}
\usepackage{dsfont}
\usepackage{hyperref}
\usepackage[table]{xcolor}
\usepackage{tabularx}
\usepackage{rotating}

\begin{document}
\title{Partially Blinded Unlearning: Class Unlearning for Deep Networks a Bayesian Perspective}
%

\author{Subhodip Panda \inst{1} \and
Shashwat Sourav \inst{2} \and
Prathosh A.P. \inst{1}}
%
\institute{Indian Institute of Science, Bengaluru, India \and
Indian Institutes of Science Education and Research, Bhopal, India}
%
\maketitle              
\begin{abstract}
In order to adhere to regulatory standards governing individual data privacy and safety, machine learning models must systematically eliminate information derived from specific subsets of a user's training data that can no longer be utilized. The emerging discipline of Machine Unlearning has arisen as a pivotal area of research, facilitating the process of selectively discarding information designated to specific sets or classes of data from a pre-trained model, thereby eliminating the necessity for extensive retraining from scratch. The principal aim of this study is to formulate a methodology tailored for the purposeful elimination of information linked to a specific class of data from a pre-trained classification network. This intentional removal is crafted to degrade the model's performance specifically concerning the unlearned data class while concurrently minimizing any detrimental impacts on the model's performance in other classes. To achieve this goal, we frame the class unlearning problem from a Bayesian perspective, which yields a loss function that minimizes the log-likelihood associated with the unlearned data with a stability regularization in parameter space. This stability regularization incorporates Mohalanobis distance with respect to the Fisher Information matrix and $l_2$ distance from the pre-trained model parameters. Our novel approach, termed \textbf{Partially-Blinded Unlearning (PBU)}, surpasses existing state-of-the-art class unlearning methods, demonstrating superior effectiveness. Notably, PBU achieves this efficacy without requiring awareness of the entire training dataset but only to the unlearned data points, marking a distinctive feature of its performance.

\end{abstract}
\section{Introduction}
In recent years, there has been a notable surge in the extensive training of Machine Learning (ML) and Deep Learning (DL) models, utilizing substantial volumes of user data across diverse applications. Consequently, the integration of users' public and private data has become deeply embedded within these models. This development prompts a critical inquiry: \textit{How can individuals be afforded flexible control over the utilization of their personal data by these models?} To address this question, contemporary regulatory frameworks concerning data privacy and protection, exemplified by the California Consumer Privacy Act (CCPA)~\cite{ccpa} and the European Union's General Data Protection Regulation (GDPR)~\cite{gdpr}, mandate organizations to implement rigorous controls on models.  Specifically, these regulations compel entities not only to expunge personal information from databases but also to effectuate the deletion of such data from the trained models upon the users' explicit request. One rudimentary approach to tackle this issue involves fully retraining a machine-learning model using the remaining data after removing the data intended for unlearning. Nevertheless, this methodology proves impractical, given its significant demands on time and space resources, especially when dealing with large remaining datasets, a common occurrence in practical scenarios. Hence, the process of unlearning persists as a challenging endeavor.

The framework of Machine Unlearning~\cite{unlearningsurvey1,unlearningsurvey2} endeavors to tackle the aforementioned challenges. Specifically, machine unlearning pertains to the task of either forgetting the learned information \cite{rmwhatforget, forget2, forget3, forget4, forget5, forget6, unlearning4, unlearning8} or erasing the influence~\cite{influenceremoval1,influenceremoval2,influenceremoval3,influenceremoval4,influenceremoval5,influenceremoval6} of a specific subset of training dataset from a learned model, in response to a user's request. Contemporary methodologies for machine unlearning within classification networks primarily revolve around the endeavor of unlearning a small subset or the entirety of data points associated with a specific class, whether it be a singular or multiple classes. This specific process is commonly referred to as \textit{Class Unlearning.} This task poses an inherent challenge as it requires the targeted `unlearning' of a particular class within the training data while avoiding adverse effects on the previously acquired knowledge from other classes of data points. In essence, the unlearning process introduces the risk of Catastrophic Forgetting, as evidenced by prior studies~\cite{unlearning4,unlearning5,unlearning8}. This risk may lead to a significant deterioration in the overall performance of the model, particularly in classes other than the one targeted for unlearning. To address these challenges, contemporary methods~\cite{bad-teaching, tarun2023fast} in class unlearning often assume access to the entire dataset and employ strategies to restore the model's performance on other data classes. In some cases~\cite{influenceremoval1, influenceremoval3}, these methods rely on stringent assumptions over the training procedure only applicable to small models.

Our study addresses the challenge of class unlearning within a pre-trained classification network. Inspired by the work of~\cite{bayesianunlearning}, we adopt a Bayesian perspective, framing the issue as the unlearning of specific data points. The theoretical formulation yields a loss function with the objective of minimizing the log-likelihood associated with the unlearned class data, incorporating a stability regularization term. Utilizing the second-order Taylor approximation and the assumption of parameter Gaussianity, the stability regularization can be decomposed into two main components. The first component centers on the Mahalanobis distance, computed relative to the Fisher Information matrix of the initial parameters. Meanwhile, the second component relates to the $l_2$ distance, quantifying the disparity between the unlearned model and the initial model.
Notably, this formulation allows for a natural interpretation of a trade-off: fully unlearning the target data class versus retaining some knowledge of other class data encoded in the initial parameters. The latter mitigates the risk of catastrophic unlearning induced by the former. Furthermore, our approach distinguishes itself by not demanding access to the complete dataset, a requirement prevalent in many contemporary methods~\cite{tarun2023fast, bad-teaching, graves2021amnesiac}. Instead, it only necessitates access to the unlearned data to yield results. In essence, it remains oblivious to the retaining data classes, earning its nomenclature as Partially-Blinded Unlearning. Our contributions can be summarized as follows:

\begin{itemize}
\item We provide a theoretical formulation for Class Unlearning and propose a methodology applicable to unlearning specific classes in deep neural networks.
\item Our method exhibits superiority over concurrent class unlearning methods~\cite{bad-teaching, tarun2023fast} by not mandating access to the entire dataset, only requiring partial access to the unlearned data. This feature proves advantageous in more restricted experimental settings.
\item Our method's single-step unlearning process contrasts with two-step approaches used by some contemporary methods~\cite{tarun2023fast}, showcasing superior computational efficiency and simplicity
\item We validate our approach using ResNet-18, ResNet-34, ResNet-50, and All-CNN models across four benchmarking vision datasets, demonstrating its generalizability across different models and datasets. Our method consistently outperforms many state-of-the-art class unlearning methods.
\end{itemize}

\section{Related Works}
\subsection{Machine Unlearning}
Machine unlearning, as defined by the literature~\cite{unlearningsurvey1,unlearningsurvey2}, involves intentionally discarding specific acquired knowledge or erasing the impact of particular subsets of training data from a trained model. The naive approach to unlearning is to remove the undesired data subset from the training dataset and then retrain the model from the beginning. However, this method becomes computationally impractical when unlearning requests are made iteratively for individual data points. The concept of recursively `unlearning', which is removing information of a single data point in an online manner (also known as decremental learning) for the SVM algorithm, was introduced in a previous study~\cite{unlearning1}. However, the efficiency of these algorithms diminishes when dealing with multiple data points being added or removed, as they need to be applied individually. To address this limitation, a more efficient SVM training algorithm was introduced by~\cite{unlearning3}, capable of updating an SVM model when multiple data points are added or removed simultaneously. Subsequently, inspired by user privacy concerns,~\cite{forget4} developed efficient methods for deleting data from certain statistical query algorithms, coining the term ``machine unlearning''. Unfortunately, these techniques are primarily suited for structured problems and do not extend well to complex machine learning algorithms like k-means clustering~\cite{unlearning4} or random forests~\cite{unlearning6}.

Later, unlearning algorithms were proposed for the k-means clustering problem, introducing effective data unlearning criteria applicable to randomized algorithms based on statistical indistinguishability. Building upon this criterion, machine unlearning methods are broadly categorized into two main types: exact unlearning~\cite{unlearning4,unlearning6} and approximate unlearning~\cite{unlearning7}. Exact unlearning aims to completely eliminate the influence of unwanted data from the trained model, requiring precise parameter distribution matching between the unlearned and retrained models. In contrast, approximate unlearning methods only partially mitigate data influence, resulting in parameter distributions closely resembling the retrained model, albeit with minor multiplicative and additive adjustments. To achieve exact unlearning,~\cite{influenceremoval} proposed a technique employing subtracting the cached gradients of the unlearning data, offering computational efficiency while increasing memory usage. Also, this method needs awareness of the unlearning data during the training phase which is quite restrictive. More sophisticated methods~\cite{influenceremoval2,influenceremoval3} have suggested using influence functions for this purpose. However, these approaches are computationally demanding due to the necessity of Hessian inversion techniques and are limited to small convex models.

To broaden the applicability of unlearning techniques to non-convex models such as deep neural networks,~\cite{unlearning8} introduced a scrubbing mechanism centered on the Fisher Information matrix. This mechanism is designed for unlearning a specific class of data, with a particular emphasis on the task of class unlearning. Despite the possibility of achieving zero accuracy on unlearned class data,~\cite{tarun2023fast} demonstrates that this scrubbing mechanism often yields poor accuracy on data from retaining classes. For further improvement,~\cite{bad-teaching} introduced a method based on the knowledge distillation approach, employing two teacher networks. One is identified as a proficient teacher, trained on the entire dataset, while the other is an unskilled teacher initialized randomly. The methodology involves fine-tuning a student network, and adjusting its parameters so that its output aligns with the proficient teacher for the retaining data class and the unskilled teacher for the unlearning data class. Despite achieving state-of-the-art results, this method mandates access to the complete training dataset, posing a significant restriction. Our current method (PBU) is close to the works of~\cite{unlearning8} as it also incorporates the Fisher Information Matrix of the initial parameters but produces superior to comparable performance. 

\subsection{Continual Learning}
Continual learning constitutes an active area of research, to devise methods that enable the model to adapt to future tasks while preserving its performance on previously acquired tasks. Even though the objective of Continual Learning and Machine Unlearning is orthogonal many techniques of continual learning can be applied to the task of unlearning~\cite{mu-cl,unlearning9}. Especially the task of incrementally learning using the Elastic Weight Consolidation (EWC) regularization~\cite{EWC, cont-ewc} is closely related to our method of unlearning. If the Fisher Information matrix becomes diagonal for our case the stability regularizer resolves to EWC regularization technique. Additionally, one has the flexibility to explore enhancements beyond EWC, such as online EWC~\cite{cont-ewc}, or other regularization-based techniques grounded in Bayesian learning principles~\cite{var-cont-learning, cont-func-regul, gen-var-cont }. 

\section{Methodology}

\subsection{Unlearning Problem Formulation}
Consider a particular parameter in the Parameter Space $\Theta \subseteq \mathds{R}^m$ is denoted as $\theta$. Now the pre-trained classification model denoted as $f_{\theta^*}$ with initial parameters $\theta^* \in \Theta$.  This classifier undergoes training using a dataset $\mathcal{D} = \{(x_i, y_i)\}_{i=1}^{|\mathcal{D}|}$, where $(x_i, y_i) \overset{iid}{\sim} P_{XY}(x, y)$. Here $x_i$ signifies the feature vector belonging to the feature space $\mathcal{X} \subseteq \mathds{R}^d$, and $y_i$ represents the corresponding label or class belonging to the label space $\mathcal{Y} = \{0, 1, 2, \ldots, C-1\}$, where $C$ stands for the total number of classes in the dataset. In response to a user-specified request, indicating a particular class or classes of data points $s_n \in \mathcal{Y}$ that the model needs to unlearn, we gain access to the unlearning samples of that specific class denoted as $\mathcal{S}_n = \{(x_i,y_i): y_i = s_n \}$. The objective of unlearning is to determine a parameter $\theta^u$ for the unlearned model $f_{\theta^u}$ that closely aligns with the performance of the retrained model $f_{\theta^p}$, which is trained on samples $\mathcal{S}_p = \mathcal{D} \setminus \mathcal{S}_n$. In essence, given that the retrained model achieves low accuracy on unlearned class samples and high accuracy on retaining class samples, the task involves adjusting the initial model parameter $\theta^*$ to $\theta^u$ such that the accuracy of $f_{\theta^u}$ on unlearned class samples $\mathcal{S}_n$ decrease, while maintaining similar performance as the retrained model on retained samples $\mathcal{S}_p$. It's crucial to highlight that we work under a more restrictive condition where access to $\mathcal{S}_p$ is prohibited, rendering us unaware of the entire dataset $\mathcal{D} = \mathcal{S}_p \cup \mathcal{S}_n$. Under this constraint of restricted access, retraining the model is infeasible in our scenario.

\subsection{Unlearning Problem: Bayesian view}
   Given $\theta^*$ and $\mathcal{S}_n$, the unlearning objective is to find the unlearned parameter $\theta^u$ that closely resembles the retrained parameter $\theta^p$ expressed via a maximum a posteriori estimate given by $\theta^p = \arg \max_{\theta} P(\theta|\mathcal{S}_p)$, which is expanded as below:
\begin{align}
    \theta^p &= \arg \max_{\theta} \log P(\theta|\mathcal{S}_p) \\
    &= \arg \max_{\theta} \log P(\mathcal{S}_p|\theta) + \log P(\theta) - \log P(\mathcal{S}_p) \label{eq-2} \\
    &= \arg \max_{\theta} \log P(\mathcal{S}_p|\theta) + \log P(\theta) - K_1 \label{eq-3}
\end{align}

Eq.-\ref{eq-2} is obtained via Bayes Rule and the term $\log P(\mathcal{S}_p)$ is replaced by a constant $K_1$ since it is independent of $\theta$, to obtain Eq.\ref{eq-3}. Now from the pre-trained model on the whole dataset $\mathcal{D}$, we have the following:
\begin{align}
    \log P(\theta|\mathcal{D}) &= \log P(\theta|\mathcal{S}_p, \mathcal{S}_n) \\
    &= \log P(\mathcal{S}_p, \mathcal{S}_n | \theta) + \log P(\theta) - K_2 \\
    &= \log P(\mathcal{S}_p | \theta) + \log P(\mathcal{S}_n|\theta) + \log P(\theta) - K_2
    \label{eq-6}
\end{align}

Now using the Eq.-\ref{eq-6} in Eq.-\ref{eq-3}, we get the following:

\begin{align}
    \theta^p &= \arg \max_{\theta} \log P(\theta|\mathcal{D}) - \log P(\mathcal{S}_n|\theta) + K_2 - K_1 \\
     &\equiv \arg \max_{\theta} \log P(\theta|\mathcal{D}) - \log P(\mathcal{S}_n|\theta)  \\
     &= \arg \max_{\theta} \mathcal{L}(\theta, \mathcal{D}, \mathcal{S}_n) \label{eq-9} 
\end{align}

\subsection{Proposed Method: Overview}
\begin{figure*}[]
    \centering
    \includegraphics[width=\textwidth]{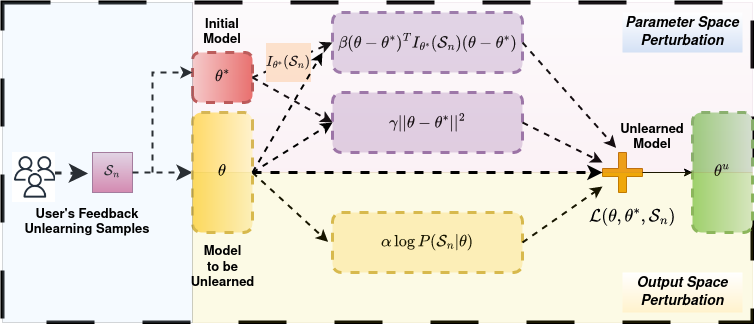}
    \caption{\textbf{Partially Blinded Unlearning (PBU) Method:} Given user-identified samples to be unlearned ($\mathcal{S}_n$), our unlearning method employs a two component perturbation technique, indicated by a loss function comprising three terms: the first term(shown in the bottom half) represents the perturbation in the output space, aiming to minimize the log-likelihood associated with the unlearned class while the last two terms correspond to perturbations in the parameter space (shown in the upper half), including the Mahalanobis Distance with respect to the Fisher Information matrix and the $l_2$ distance.}
    \label{fig:PBU-method}
\end{figure*}  

The task of finding the unlearned parameter $\theta^u$, that approximates $\theta^p$ in Eq. \ref{eq-9} demands access to the complete dataset $\mathcal{D}$, which is infeasible. Given that only the unlearning data $\mathcal{S}_n$ is accessible, we reformulate the objective (maximizing $\mathcal{L}(\theta, \mathcal{D}, \mathcal{S}_n)$) by constructing an upper bound on it\footnote{While maximization of a lower bound of the objective is desirable, we resort to an upper bound optimization since construction of a lower bound is intractable in the current problem and show that it leads to better performance empirically.}. This results in the minimization of a loss function $\mathcal{L}(\theta, \theta^*, \mathcal{S}_n)$(shown in Remark-\ref{remark-1}), as given in Eq.\ref{eq-10}.

\begin{equation}
    \mathcal{L}(\theta, \theta^*, \mathcal{S}_n) = \alpha \log P(\mathcal{S}_n|\theta) + \beta (\theta - \theta^*)^T I_{\theta^*}(\mathcal{S}_n)(\theta - \theta^*) + \gamma ||\theta - \theta^*||^2 
    \label{eq-10}
\end{equation} Here, $\alpha, \beta, \gamma$ are hyper-parameters and $I_{\theta^*}(\mathcal{S}_n)$ is the Fisher Information Matrix of the initial parameters corresponding to the unlearning data class. The overall methodology, illustrated in Figure-\ref{fig:PBU-method}, entails perturbing the initial parameter $\theta^*$ to the unlearned parameter $\theta^u$ using the loss function described in Equation-\ref{eq-10}. The loss function comprises two components: the first component involves perturbing the parameters by minimizing the log-likelihood associated with the unlearning class data, while the second component involves perturbing parameters by incorporating stability regularization in the parameter space. 
This stability regularization encompasses the Mahalanobis distance between the initial parameter and the unlearned parameter with respect to the Fisher Information matrix corresponding to the negative class, along with the $l_2$ distance between the initial parameter and the unlearned parameter. Through this methodology, our method strives to strike a balance between erasing specific class information and retaining overall model performance and robustness. In our formulation, the dependence on $\mathcal{D}$ is encapsulated in the initial parameter $\theta^*$. This characteristic provides an advantage to our method, as it only requires access to the unlearning class data points, making it \textbf{\textit{`partially blind'}} to the whole dataset. Also, Our PBU algorithm-\ref{alg1} is a single step method and does not require repairing as suggested in many of the algorithms~\cite{tarun2023fast, bad-teaching}. In the subsequent section, we provide a theoretical exposition detailing the construction of the loss function utilized in our methodology.

\begin{algorithm}
    \caption{Partially Blinded Unlearning(PBU)}\label{alg1}
    \textbf{Input}: Unlearning Data Classes: $\mathcal{S}_n$, Initial Paramter: $\theta^*$, hyper-parameters: $\alpha, \beta, \gamma$
    \begin{algorithmic}
        \STATE{Initialize: $\theta \gets \theta^*$} 
        \STATE{Calculate Fisher Information Matrix: $I_{\theta^*}(\mathcal{S}_n)$}
        \FOR{$t \leq T_{ul}$}
            \STATE{Calculate $\mathcal{L}(\theta, \theta^*, \mathcal{S}_n) = \alpha \log P(\mathcal{S}_n|\theta) + \beta (\theta - \theta^*)^T I_{\theta^*}(\mathcal{S}_n)(\theta - \theta^*) + \gamma ||\theta - \theta^*||^2$}
            \STATE{$\theta^{t+1} \gets \theta^{t} - \eta \nabla_{\theta}\mathcal{L}(\theta, \theta^*, \mathcal{S}_n)$}
        \ENDFOR \\
    \textbf{Output}: $\theta^{T_{ul}}$
    \end{algorithmic}
\end{algorithm}


\subsection{Construction of Loss Function: Theoretical Outlook \label{sec 3.4}}


\begin{definition}
\textbf{(Fisher Information Matrix):} $\forall \theta \in \Theta$ the Fisher Information Matrix($I_{\theta}(\mathcal{D})$) for the model on the whole dataset $\mathcal{D}$ is defined as follows:
\begin{equation}
  I_{\theta}(\mathcal{D}) = \mathds{E}_{P_{\theta}(\mathcal{D})}\left[\nabla_{\theta} \log P_{\theta}(\mathcal{D})\nabla_{\theta} \log P_{\theta}(\mathcal{D})^T\right]
\end{equation}
\end{definition}

\textbf{Regularity Conditions:} Now the above definition can further be simplified if we assume certain regularity conditions as follows:
\begin{enumerate}
    \item \label{reg-1}$\Theta$ be an open set in $\mathds{R}^m$
    \item \label{reg-2}The partial derivatives $\frac{\partial P_{\theta}(\mathcal{D})}{\partial \theta_i} \text{and} \frac{\partial^2 P_{\theta}(\mathcal{D})}{\partial \theta^2_i} ;\forall i \in \{1,2,\ldots,m\}$ exists and is finite $\forall \theta \in \Theta$ and $\forall x \in \mathcal{X}$
    \item \label{reg-3}$\int_{\mathcal{X}} \nabla_{\theta} P_{\theta}(\mathcal{D}) d\mu(\mathcal{D}) = \nabla_{\theta} \int_{\mathcal{X}} P_{\theta}(\mathcal{D}) d\mu(\mathcal{D}) = 0$
    \item \label{reg-4}$\int_{\mathcal{X}}\nabla_{\theta}^2 P_{\theta}(\mathcal{D})  d\mu(\mathcal{D}) = \nabla_{\theta}^2 \int_{\mathcal{X}} P_{\theta}(\mathcal{D})  d\mu(\mathcal{D}) = 0$
\end{enumerate}

\begin{theorem}\footnote{The proof is can be found in the cited text.}
    \textbf{(Book chapters~\cite{moulin-veravalli}Lemma-13.1)} If the regularity conditions-(\ref{reg-1})-(\ref{reg-4}) hold then  $\forall \theta \in \Theta$ the Fisher Information matrix can be written in the following form
    \begin{equation}
        I_{\theta}(\mathcal{D}) = \mathds{E}_{P_{\theta}(\mathcal{D})}\left[-\nabla_{\theta}^2 \log P_{\theta}(\mathcal{D})\right]
    \end{equation}
\end{theorem}

\begin{lemma}
    If the regularity conditions-(\ref{reg-1})-(\ref{reg-4}) hold then for the pretrained initial parameter $\theta^*$
    \begin{equation}
        I_{\theta^*}(\mathcal{D}) \geq I_{\theta^*}(\mathcal{S}_n) 
    \end{equation}
\end{lemma}
\begin{proof}
    Under the regularity conditions-(\ref{reg-1})-(\ref{reg-4}) by Theorem-1 we can write
\begin{align*}
    I_{\theta^*}(\mathcal{D}) &= \mathds{E}_{P_{\theta^*}(\mathcal{D})}\left[-\nabla_{\theta}^2 \log P_{\theta^*}(\mathcal{D})\right] \\
    &= \mathds{E}_{P_{\theta^*}(\mathcal{D})}\left[-\nabla_{\theta}^2 \log P_{\theta^*}(\mathcal{S}_p,\mathcal{S}_n)\right] \\
    &\stackrel{(a)}{=} \mathds{E}_{P_{\theta^*}(\mathcal{D})}\left[-\nabla_{\theta}^2 \log P_{\theta^*}(\mathcal{S}_p) - \nabla_{\theta}^2 \log P_{\theta^*}(\mathcal{S}_n) \right] \\
    &\stackrel{(b)}{=} \mathds{E}_{P_{\theta^*}(\mathcal{S}_p)}\left[-\nabla_{\theta}^2 \log P_{\theta^*}(\mathcal{S}_p)\right] + \mathds{E}_{P_{\theta^*}(\mathcal{S}_n)} \left[-\nabla_{\theta}^2 \log P_{\theta^*}(\mathcal{S}_n) \right] \\
    &= I_{\theta^*}(\mathcal{S}_p) + I_{\theta^*}(\mathcal{S}_n) \\
    &\geq I_{\theta^*}(\mathcal{S}_n) \quad \left(I_{\theta^*}(\mathcal{S}_p) \geq 0 \right)
\end{align*}
\end{proof}

In the above proof, (a) holds because, via i.i.d assumption, $\forall \theta \in \Theta, P_{\theta}(\mathcal{S}_p,\mathcal{S}_n) = P_{\theta}(\mathcal{S}_p) P_{\theta}(\mathcal{S}_n)$. Also (b) is true by the linearity of Expectation and Marginalization.

\begin{lemma}
\label{lemma-2}
(\textbf{~\cite{influenceremoval} Section-4.2, ~\cite{influenceremoval4,unlearning9}}) If $\theta^*$ is the MAP estimate of $P(\theta|\mathcal{D})$ then for some constant $K_3$ independent of the parameter $\theta$ the following approximation holds.
    \begin{align*}
        \log P(\theta|\mathcal{D}) \approx \frac{1}{2} (\theta - \theta^*)^T \nabla_{\theta}^2 \log P(\theta^*|\mathcal{D}) (\theta - \theta^*) + K_3
    \end{align*}
\end{lemma}

\begin{proof}
    As $\theta^*$ is the MAP estimate of $\log P(\theta|\mathcal{D})$, 
    \begin{align}
        \nabla_{\theta} \log P(\theta^*|\mathcal{D}) = 0
    \end{align}
    Now using second-order Taylor-Series approximation of $\log P(\theta|\mathcal{D})$ around $\theta^*$ gives:
    \begin{align*}
    \log P(\theta|\mathcal{D}) &\approx \log P(\theta^*|\mathcal{D}) + (\theta - \theta^*)^T \nabla_{\theta} \log P(\theta^*|\mathcal{D}) \\
    &\quad+ \frac{1}{2!} (\theta - \theta^*)^T \nabla_{\theta}^2 \log P(\theta^*|\mathcal{D}) (\theta - \theta^*) \\
    &= K_3 + 0 + \frac{1}{2} (\theta - \theta^*)^T \nabla_{\theta}^2 \log P(\theta^*|\mathcal{D}) (\theta - \theta^*)
\end{align*}
\end{proof}

\begin{theorem}
\label{theorem-2}
   If the above regularity conditions-(\ref{reg-1})-(\ref{reg-4}) hold and if the pre-trained initial parameter $\theta^*$ is the MAP estimate of $P(\theta|\mathcal{D})$ with prior parameter distribution $P(\theta) = \mathcal{N}(0,\lambda I)$ with some variance parameter $\lambda \geq 0$ then 
    \begin{align*}
        \mathcal{L}(\theta, \mathcal{D}, \mathcal{S}_n) \leq - \log P(\mathcal{S}_n|\theta) - \frac{N}{2}(\theta - \theta^*)^T I_{\theta^*}(\mathcal{S}_n)(\theta - \theta^*) - \frac{\lambda}{2}||\theta - \theta^*||^2 + K_3
    \end{align*}
\end{theorem}

\begin{proof}
By Bayes rule, we can write
\begin{align}
    P(\theta|\mathcal{D}) &= \frac{P(\mathcal{D}|\theta) P(\theta)}{P(\mathcal{D})} \\
    \implies \log P(\theta|\mathcal{D}) &= \log P(\mathcal{D}|\theta) + \log P(\theta) - \log \mathcal{D} \\
    \implies \nabla^2_{\theta} \log P(\theta|\mathcal{D}) &= \nabla^2_{\theta} \log P(\mathcal{D}|\theta) + \nabla^2_{\theta} \log P(\theta) \\
    \implies \nabla^2_{\theta} \log P(\theta^*|\mathcal{D}) &= \nabla^2_{\theta} \log P(\mathcal{D}|\theta^*) - \lambda I \\
    &\stackrel{(c)}{\approx} N \mathds{E}_{P_{\theta^*}(\mathcal{D})} \left[ \nabla^2_{\theta} \log P_{\theta^*}(\mathcal{D}) \right] - \lambda I \label{eq-17}\\
    &\stackrel{(d)}{=} - N I_{\theta^*}(\mathcal{D}) - \lambda I \label{eq-18}\\
    &\stackrel{(e)}{\leq} - N  I_{\theta^*}(\mathcal{S}_n) - \lambda I \label{eq-19}
\end{align}
 By the assumption, if $P(\theta) = \mathcal{N}(0,\lambda I)$ then $\nabla^2_{\theta} \log P(\theta) = - \lambda I$. So by the Monte-Carlo approximation of the Fisher Information Matrix $(c)$ hold implying the equation-\ref{eq-17}. Also, by the above regularity conditions-(\ref{reg-1})-(\ref{reg-4}) Theorem 1 is satisfied and $(d)$ holds. Finally applying the Lemma-1 we get the inEq.-\ref{eq-19}. 
 
Now, from equation-\ref{eq-9} the expression $\mathcal{L}(\theta, \mathcal{D}, \mathcal{S}_n)$ is defined as follows:
    \begin{align*}
        \mathcal{L}(\theta, \mathcal{D}, \mathcal{S}_n) &= \log P(\theta|\mathcal{D}) - \log P(\mathcal{S}_n|\theta) \\
        &\stackrel{(f)}{\approx} - \log P(\mathcal{S}_n|\theta) + \frac{1}{2} (\theta - \theta^*)^T \nabla_{\theta}^2 \log P(\theta^*|\mathcal{D}) (\theta - \theta^*) + K_3 \\
        &\stackrel{(g)}{\leq} - \log P(\mathcal{S}_n|\theta) - \frac{N}{2}(\theta - \theta^*)^T I_{\theta^*}(\mathcal{S}_n)(\theta - \theta^*) - \frac{\lambda}{2}||\theta - \theta^*||^2 + K_3
    \end{align*}
\end{proof}
The implications $(f)$ and $(g)$ hold via Lemma-2 and inEq.-\ref{eq-19} respectively.

\begin{remark}
\label{remark-1}
    Now using above Theorem-2 in Eq.-\ref{eq-9} we get the loss function as follows:
Our final objective to find $\theta^p$ is as follows 
\begin{align}
    \theta^p &= \arg \max_{\theta} \mathcal{L}(\theta, \mathcal{D}, \mathcal{S}_n) \\
    &\leq \arg \max_{\theta} - \log P(\mathcal{S}_n|\theta) - \frac{N}{2}(\theta - \theta^*)^T I_{\theta^*}(\mathcal{S}_n)(\theta - \theta^*) - \frac{\lambda}{2}||\theta - \theta^*||^2 + K_3 \\
    &\equiv \arg \min_{\theta}  \log P(\mathcal{S}_n|\theta) + \frac{N}{2}(\theta - \theta^*)^T I_{\theta^*}(\mathcal{S}_n)(\theta - \theta^*) + \frac{\lambda}{2}||\theta - \theta^*||^2 \label{eq-22} \\
    &\approx \arg \min_{\theta} \alpha \log P(\mathcal{S}_n|\theta) + \beta (\theta - \theta^*)^T I_{\theta^*}(\mathcal{S}_n)(\theta - \theta^*) + \gamma ||\theta - \theta^*||^2
\end{align}
\end{remark}

\section{Experiments and Results}

\subsection{Datasets, Models and Unlearned Classes}

For Class Unlearning objective, we have used four classification models: ResNet18~\cite{he2016deep}, ResNet34~\cite{he2016deep}, ResNet50~\cite{he2016deep} and AllCNN~\cite{challa2019multi} models for learning and unlearning tasks. Further to evaluate the applicability of our method we take four datasets, i.e., MNIST~\cite{lecun1998gradient}, CIFAR10, CIFAR100 ~\cite{krizhevsky2009learning}, and Food101~\cite{bossard2014food} dataset. For MNIST, we unlearned classes 2,6 and 8 in Resnet-18,Resnet-34 and All-CNN models.  Similarly, in the case of CIFAR10, we unlearned classes 1,3 and 5 for the same set of models as in MNIST. For CIFAR100, we used ResNet-18, ResNet-34, and ResNet-50 to unlearn classes 1,3 and 8. For the Food101 dataset, we used the same set of models as used in CIFAR100, but we unlearned classes 1,3 and 8 while using ResNet18, classes 10,30 and 50 while using ResNet50, and classes 20,40 and 60 to perform unlearning using ResNet34.


\subsection{Implementation Details}

\subsubsection{Initial Training, Unlearning, and Baselines:} Initially, we trained all models (referred to as the initial model) using the entire dataset. Subsequently, we partitioned the test data into two segments: one containing the class slated for unlearning and the other encompassing the remaining classes. We assessed the accuracy of the initial model on both partitions of the test data. In our approach, we solely train the unlearned model using the unlearning class and employ a proposed loss function in eq-\ref{eq-10} to facilitate unlearning of the desired class. Additional specifics regarding the Initial Training and Unlearning methodologies are provided in the Appendix. To evaluate our method, we included four baseline techniques. These are as follows:
\begin{itemize}
    \item \textbf{Retraining:} We trained the model anew using data excluding the class designated for unlearning. The retrained model exhibited 0\% accuracy on the unlearned class, in contrast to the initial model's significant accuracy on the same class, while maintaining similar accuracy on the retained classes.
    \item \textbf{Fine-Tuning:} Rather than retraining from scratch, we initialized the model from the initial pre-trained checkpoint and fine-tuned it using the retained class data points. 
    \item \textbf{Fast-Effective:} We conducted a comparative analysis between our model and the Fast Yet Effective approach proposed by ~\cite{tarun2023fast}. In their method, the authors maximize noise by employing a noise matrix to manipulate the weights of the initial model for unlearning through impair-repair steps, where weights are manipulated in the impair step and stabilized on the retained classes in the subsequent repair step. 
    \item \textbf{Bad Teaching:} Additionally, our method was compared with the bad teaching approach proposed by~\cite{bad-teaching}. In this method, a competent teacher, an incompetent teacher, and a student model are involved, where the student model attempts to unlearn classes specified by the user based on the information received from both teachers. 
\end{itemize}

\subsection{Evaluation Metrics}

In previous works ~\cite{golatkar2020eternal,golatkar2020forgetting,golatkar2021mixed,graves2021amnesiac},unlearning metrics have been introduced to measure the performance of the unlearning algorithm. These metrics have been developed to measure the amount of information left in the network corresponding to the forget class. We mainly use the following metrics :

\begin{itemize}
    \item \textbf{Accuracy on Forget and Retain Class:} Following unlearning, the model's accuracy on the forgotten class, denoted as $A_{D_{f}}$, is expected to be significantly low (approaching zero), while the accuracy on the remaining classes, represented as $A_{D_{r}}$, ideally should closely resemble the performance of the retrained model.
    \item \textbf{Membership Inference Attack(MIA) Accuracy:} This metric signifies the model's resilience against membership inference attacks, aiming to prevent the extraction of information regarding a user's presence in the dataset. Ideally, the Membership Inference Attack (MIA) accuracy should remain below 0.5, indicating that any unlearning mechanism should not render the model more susceptible to membership attacks than random chance.
    \item \textbf{Unlearning Time:} The unlearning method is expected to be computationally lightweight, incurring low computational cost. To assess computational efficiency, the method should yield satisfactory results within a limited number of epochs. Therefore, the runtime, measured in terms of epochs, serves as a crucial metric for evaluating various unlearning methods. 
\end{itemize}

\subsection{Unlearning Results}

\begin{table*}[!t]
\centering
\resizebox{\textwidth}{!}{%
\begin{tabular}{c|cc|cc|cc cc cc cc|cc}
\toprule
\multirow{2}{*}{\textbf{Dataset}} & \multirow{2}{*}{\textbf{Models}} & \multirow{2}{*}{\textbf{Classes}} & \multicolumn{2}{c|}{\textbf{Initial Training}} & \multicolumn{2}{c}{\textbf{Re-training}} & \multicolumn{2}{c}{\textbf{Fine-tuning}} & \multicolumn{2}{c}{\textbf{Fast-Effective}~\cite{tarun2023fast}} & \multicolumn{2}{c|}{\textbf{Bad Teaching}~\cite{bad-teaching}} & \multicolumn{2}{c}{\textbf{Our Method(PBU)}} \\

& & & $A_{D_f}$ & $A_{D_r}$ & $A_{D_f}$ & $A_{D_r}$ & $A_{D_f}$ & $A_{D_r}$ & $A_{D_f}$ & $A_{D_r}$ & $A_{D_f}$ & $A_{D_r}$ & $A_{D_f}$ & $A_{D_r}$ \\
\hline

\multirow{9}{*}{\begin{turn}{90}
    \textbf{CIFAR-100}
\end{turn}} & \multirow{3}{*}{\textbf{Resnet-18}} & Class-1 & 85.67$\pm$2.08 & 76.09$\pm$0.98 & 0$\pm$0 & 74.9$\pm$1.09 & 0$\pm$0 & 74.53$\pm$0.74 & 0$\pm$0 & 64.97$\pm$0.54 & 0$\pm$0 & 63.35$\pm$0.03 & \textbf{0$\pm$0.58} & \textbf{70.54$\pm$0.28} \\
&  & Class-3 & 60.33$\pm$4.04 & 76.22$\pm$1.2 & 0$\pm$0 & 75.19$\pm$0.09 & 0$\pm$0 & 74.94$\pm$0.23 & 0$\pm$0 & 62.34$\pm$1.13 & 0.67$\pm$0.58 & 65.45$\pm$0 & \textbf{1.5$\pm$1.73} & \textbf{70.66$\pm$0.35} \\
& & Class-8 & 94.33$\pm$4.04 & 75.88$\pm$1.17 & 0$\pm$0 & 74.81$\pm$0.64 & 0$\pm$0 & 74.6$\pm$0.59 & \textbf{0$\pm$0} & \textbf{63.82$\pm$0.22} & 0$\pm$0 & 62.79$\pm$0.54 & 3$\pm$0.58 & 67.45$\pm$0.77 \\

\cmidrule{2-15}

& \multirow{3}{*}{\textbf{Resnet-50}} & Class-1 & 86.33$\pm$7.23 & 75.87$\pm$0.31 & 0$\pm$0 & 75.2$\pm$0.34 & 0$\pm$0 & 69.98$\pm$1.24 & 0$\pm$0 & 54.61$\pm$0.21 & 0.87$\pm$0.23 & 68.06$\pm$0.14 & \textbf{0$\pm$1.15} & \textbf{70.51$\pm$0.18} \\
& & Class-3 & 56.67$\pm$11.72 & 76.17$\pm$0.41 & 0$\pm$0 & 74.12$\pm$0.93 & 0$\pm$0 & 69.25$\pm$0.36 & 0$\pm$0 & 59.43$\pm$0.56 & 0$\pm$0 & 70.35$\pm$1.19 & \textbf{0.5$\pm$0.58} & \textbf{71.85$\pm$0.91} \\
& & Class-8 & 92.33$\pm$2.89 & 75.81$\pm$0.34 & 0$\pm$0 & 74.31$\pm$0.83 & 0$\pm$0 & 68.28$\pm$0.66 & 0$\pm$0 & 57$\pm$0.1 & \textbf{0$\pm$0} & \textbf{65.98$\pm$0.78} & 0.5$\pm$1 & 65.51$\pm$2.05 \\

\cmidrule{2-15}

& \multirow{3}{*}{\textbf{Resnet-34}} & Class-1 & 88$\pm$2.65 & 75.17$\pm$0.8 & 0$\pm$0 & 67.61$\pm$5.26 & 0$\pm$0 & 66.74$\pm$0.91 & 0$\pm$0 & 54.89$\pm$0.21 & 0$\pm$0 & 53.83$\pm$0.93 & \textbf{0$\pm$0} & \textbf{64.31$\pm$1.21} \\
& & Class-3 & 74.67$\pm$10.02 & 75.3$\pm$0.68 & 0$\pm$0 & 63.99$\pm$1.75 & 0$\pm$0 & 68.05$\pm$0.7 & 0$\pm$0 & 55.75$\pm$1.36 & 0$\pm$0 & 51.77$\pm$0.57 & \textbf{0$\pm$0} & \textbf{68.07$\pm$1.41} \\
& & Class-8 & 89.33$\pm$0.58 & 75.15$\pm$0.78 & 0$\pm$0 & 64.3$\pm$0.76 & 0$\pm$0 & 67.46$\pm$0.84 & 0$\pm$0 & 57.23$\pm$0.53 & 0$\pm$0 & 54.07$\pm$0.46 & \textbf{0.5$\pm$0.58} &\textbf{ 63.63$\pm$2.05} \\

\midrule

\multirow{9}{*}{\begin{turn}{90}
    \textbf{FOOD-101}
\end{turn}} & \multirow{3}{*}{\textbf{Resnet-18}} & Class-1 & 76$\pm$3.46 & 74.87$\pm$1.06 & 0$\pm$0 & 73.29$\pm$0.06 & 0$\pm$0 & 72.27$\pm$2.14 & 0$\pm$0 & 57.94$\pm$0.61 & 0.26$\pm$0.23 & 58.79$\pm$0.39 & \textbf{2.53$\pm$0.46} & \textbf{67.65$\pm$0.72} \\
& & Class-3 & 83.6$\pm$3.46 & 74.79$\pm$1.13 & 0$\pm$0 & 73.03$\pm$0.62 & 0$\pm$0 & 73.81$\pm$0.01 & 0$\pm$0 & 61.1$\pm$0.98 & 0.33$\pm$0.11 & 59.54$\pm$0.29 & \textbf{2.13$\pm$1.85} & \textbf{63.77$\pm$0.62} \\
& & Class-8 & 46.93$\pm$3.93 & 75.16$\pm$1.06 & 0$\pm$0 & 73.18$\pm$0.6 & 0$\pm$0 & 73.97$\pm$0.01 & 0$\pm$0 & 67.54$\pm$0.54 & 0$\pm$0 & 61.41$\pm$0.87 & \textbf{0$\pm$0} & \textbf{68.03$\pm$0.23} \\

\cmidrule{2-15}

& \multirow{3}{*}{\textbf{Resnet-50}}& Class-10 & 67.2$\pm$5.54 & 78.18$\pm$0.01 & 0$\pm$0 & 75.1$\pm$0.23 & 0$\pm$0 & 77.3$\pm$0.77 & 0$\pm$0 & 60.83$\pm$1.4 & 0$\pm$0 & 56.07$\pm$1.08 & \textbf{0.8$\pm$0.69} & \textbf{68.34$\pm$1.24} \\
& & Class-30 & 90.8$\pm$4.16 & 77.94$\pm$0.01 & 0$\pm$0 & 74.86$\pm$0.09 & 0$\pm$0 & 77.08$\pm$0.25 & 0$\pm$0 & 62.13$\pm$0.95 & 0$\pm$0 & 52.45$\pm$0.57 & \textbf{0.27$\pm$0.46} & \textbf{65.46$\pm$0.32} \\
& & Class-50 & 59.6$\pm$4.85 & 78.26$\pm$0 & 0$\pm$0 & 74.18$\pm$0.2 & 0$\pm$0 & 77.23$\pm$0.32 & 0$\pm$0 & 63.06$\pm$0.77 & 0$\pm$0 & 55.53$\pm$0.48 & \textbf{0$\pm$0} & \textbf{69.43$\pm$0.81} \\

\cmidrule{2-15}

& \multirow{3}{*}{\textbf{Resnet-34}} & Class-20 & 76.67$\pm$16.86 & 64.88$\pm$1.03 & 0$\pm$0 & 67.04$\pm$1.52 & 0$\pm$0 & 71.49$\pm$0.74 & 0$\pm$0 & 39.46$\pm$0.73 & 1.12$\pm$0.39 & 51.42$\pm$1.53 & \textbf{0.6$\pm$0.2} & \textbf{59.75$\pm$0.36} \\
& & Class-40 & 77.2$\pm$9.01 & 64.87$\pm$1.1 & 0$\pm$0 & 66.02$\pm$0.29 & 0$\pm$0 & 72.23$\pm$1.5 & 0$\pm$0 & 47.51$\pm$0.63 & \textbf{0.39$\pm$0.08} & \textbf{51.79$\pm$0.54} & 2.13$\pm$0.23 & 51.86$\pm$0.72 \\
& & Class-60 & 71.87$\pm$3.23 & 64.92$\pm$1.22 & 0$\pm$0 & 67.48$\pm$1.52 & 0$\pm$0 & 71.28$\pm$1.09 & 0$\pm$0 & 42.67$\pm$0.51 & 3.42$\pm$0.67 & 43.51$\pm$0.54 & \textbf{0.93$\pm$0.61} & \textbf{52.44$\pm$2.17} \\

\midrule

\multirow{9}{*}{\begin{turn}{90}
    \textbf{CIFAR-10}
\end{turn}} & \multirow{3}{*}{\textbf{Resnet-18}} & Class-1 & 88.8$\pm$3 & 74.65$\pm$0.46 & 0$\pm$0 & 68.89$\pm$0.26 & 0$\pm$0 & 75.66$\pm$0.75 & 0$\pm$0 & 50.82$\pm$3.14 & 0$\pm$0 & 26.7$\pm$0 & \textbf{0.97$\pm$1} & \textbf{64.23$\pm$2} \\
& & Class-3 & 62.07$\pm$2.02 & 77.62$\pm$0.15 & 0$\pm$0 & 73.2$\pm$0.76 & 0$\pm$0 & 80.47$\pm$0.39 & \textbf{0$\pm$0} & \textbf{53.59$\pm$1.98} & 0$\pm$0 & 27.34$\pm$0 & 5.33$\pm$0 & 68.75$\pm$5 \\
& & Class-5 & 64.6$\pm$5.05 & 77.34$\pm$0.64 & 0$\pm$0 & 72.61$\pm$0.33 & 0$\pm$0 & 79.64$\pm$1.05 & \textbf{0$\pm$0} &\textbf{ 54.36$\pm$1.41} & 0$\pm$0 & 36.57$\pm$0 & 7.2$\pm$1 & 65.49$\pm$5 \\

\cmidrule{2-15}

& \multirow{3}{*}{\textbf{All-CNN}}& Class-1 & 95.33$\pm$1.15 & 83.19$\pm$0.12 & 0$\pm$0 & 83.78$\pm$0.27 & 0$\pm$0 & 83.85$\pm$0.8 & \textbf{0$\pm$0} & \textbf{54.38$\pm$2.78} & 0$\pm$0 & 17.73$\pm$0 & \textbf{2.43$\pm$1} & \textbf{71.51$\pm$2} \\
& & Class-3 & 69.07$\pm$4.51 & 86.06$\pm$0.47 & 0$\pm$0 & 86.7$\pm$0.14 & 0$\pm$0 & 87.88$\pm$0.76 & 0$\pm$0 & 53.36$\pm$1.62 & 0$\pm$0 & 34.57$\pm$0 & \textbf{0.23$\pm$0} & \textbf{81.74$\pm$1} \\
& & Class-5 & 74.53$\pm$2.31 & 85.5$\pm$0.27 & 0$\pm$0 & 86.67$\pm$0.61 & 0$\pm$0 & 86.83$\pm$0.86 & 0$\pm$0 & 58.06$\pm$0.66 & 0$\pm$0 & 27.23$\pm$0 & \textbf{0.47$\pm$1} & \textbf{75.2$\pm$1} \\

\cmidrule{2-15}

& \multirow{3}{*}{\textbf{Resnet-34}} & Class-1 & 84.43$\pm$5.93 & 75.9$\pm$1.36 & 0$\pm$0 & 76.22$\pm$0.02 & 0$\pm$0 & 75.64$\pm$1.34 & 0$\pm$0 & 52.14$\pm$1.62 & 0$\pm$0 & 18$\pm$3.46 & \textbf{0.4$\pm$0} & \textbf{65.53$\pm$0} \\
& & Class-3 & 57.27$\pm$2.11 & 77.59$\pm$1.72 & 0$\pm$0 & 79.93$\pm$0.27 & 0$\pm$0 & 80.87$\pm$0.44 & 0$\pm$0 & 56.31$\pm$4.05 & 27.38$\pm$47.43 & 7.15$\pm$0 & \textbf{1.13$\pm$1} & \textbf{72.47$\pm$1} \\
& & Class-5 & 63.87$\pm$2.37 & 76.86$\pm$1.56 & 0$\pm$0 & 79.74$\pm$1.18 & 0$\pm$0 & 79.8$\pm$0.42 & 0$\pm$0 & 59.71$\pm$1.26 & 0$\pm$0 & 18.62$\pm$1.28 & \textbf{0.23$\pm$0} & \textbf{67.82$\pm$1} \\

\midrule

\multirow{9}{*}{\begin{turn}{90}
    \textbf{MNIST}
\end{turn}} & \multirow{3}{*}{\textbf{Resnet-18}} & Class-2 & 99.48$\pm$0.31 & 99$\pm$0.16 & 0$\pm$0 & 99.19$\pm$0.23 & 0$\pm$0 & 98.46$\pm$1.32 & 0$\pm$0 & 96.22$\pm$1.01 & 0$\pm$0 & 12.56$\pm$0 & \textbf{0$\pm$0} & \textbf{97.62$\pm$0.18} \\
& & Class-6 & 99.2$\pm$0.21 & 98.98$\pm$0.17 & 0$\pm$0 & 99.04$\pm$0.52 & 0$\pm$0 & 99.33$\pm$0.08 & \textbf{0$\pm$0} & \textbf{93.63$\pm$2.16} & 0$\pm$0 & 12.48$\pm$0 & \textbf{0.53$\pm$0.92} & \textbf{94.35$\pm$2.3} \\
& & Class-8 & 98.89$\pm$1.16 & 99.03$\pm$0.14 & 0$\pm$0 & 99.17$\pm$0.31 & 0$\pm$0 & 99.17$\pm$0.25 & 0$\pm$0 & 93.71$\pm$2.07 & 0$\pm$0 & 12.48$\pm$0 & \textbf{0.1$\pm$0.17} & \textbf{96.74$\pm$1.01} \\

\cmidrule{2-15}

& \multirow{3}{*}{\textbf{All-CNN}} & Class-2 & 99.42$\pm$0.49 & 99.48$\pm$0.08 & 0$\pm$0 & 99.33$\pm$0.07 & 0$\pm$0 & 99.29$\pm$0.23 & 0$\pm$0 & 96.15$\pm$1.03 & 0$\pm$0 & 13.18$\pm$0.98 & \textbf{0$\pm$0} & \textbf{98.84$\pm$0.19} \\
& & Class-6 & 99.06$\pm$0.69 & 99.52$\pm$0.06 & 0$\pm$0 & 99.18$\pm$0.7 & 0$\pm$0 & 99.35$\pm$0.12 & 0$\pm$0 & 97.22$\pm$1 & 0$\pm$0 & 12.55$\pm$0 &\textbf{ 0$\pm$0} & \textbf{98.41$\pm$0.46} \\
& & Class-8 & 99.62$\pm$0.15 & 99.46$\pm$0.15 & 0$\pm$0 & 99.34$\pm$0.14 & 0$\pm$0 & 99.14$\pm$0.2 & \textbf{0$\pm$0} & \textbf{96.72$\pm$1.36} & 0$\pm$0 & 12.31$\pm$0 & \textbf{0.21$\pm$0.36} & \textbf{97.95$\pm$1.6} \\
\cmidrule{2-15}

& \multirow{3}{*}{\textbf{Resnet-34}} & Class-2 & 99.65$\pm$0.11 & 99.42$\pm$0.11 & 0$\pm$0 & 99.33$\pm$0.11 & 0$\pm$0 & 99.37$\pm$0.06 & 0$\pm$0 & 94.17$\pm$0.88 & \textbf{0$\pm$0} & \textbf{97.96$\pm$0.29} & \textbf{0.03$\pm$0.06} & \textbf{98.73$\pm$0.57} \\
& & Class-6 & 98.89$\pm$0.59 & 99.51$\pm$0.06 & 0$\pm$0 & 99.34$\pm$0.14 & 0$\pm$0 & 99.44$\pm$0.16 & 0$\pm$0 & 89.13$\pm$2.86 & 0$\pm$0 & 87.62$\pm$0.49 & \textbf{0$\pm$0 }& \textbf{91.09$\pm$2.9} \\
& & Class-8 & 99.73$\pm$0.21 & 99.41$\pm$0.12 & 0$\pm$0 & 99.31$\pm$0.08 & 0$\pm$0 & 99.37$\pm$0.19 & 0$\pm$0 & 94.64$\pm$1.97 & 0$\pm$0 & 96.18$\pm$0.53 & \textbf{0$\pm$0} & \textbf{98.24$\pm$0.36} \\
\bottomrule

\end{tabular}%
}
\caption{accuracy on the forgotten class: $A_{D_{f}}(\%)$ and accuracy on the remaining classes: $A_{D_{r}}(\%)$ on MNIST,  CIFAR-10, CIFAR-100 and FOOD-101 datsets for different models and unlearning-classes.}
\label{tab:performance}
\end{table*}

Experiments shown in Table-\ref{tab:performance} comprehensively detail the performance evaluation of differ ent models: Resnet-18, All-CNN, Resnet-34 and Resnet-50 across four distinct datasets, considering various class unlearning settings. Across different datasets and models, our method consistently achieves lower accuracy on the forgotten class ($A_{D_f}$) during various unlearning tasks while the high accuracy on the remaining classes ($A_{D_r}$). 

For the MNIST dataset, it is evident across various models and unlearning classes that all methods consistently attain minimal accuracy (approximately $0$) on the unlearning class ($A_{D_f}$). Notably, our proposed method exhibits superior efficacy compared to other baseline methods, particularly in terms of retained class accuracy ($A_{D_r}$), achieving a level of performance akin to retraining with higher precision. In the context of the CIFAR-10, CIFAR-100 datasets, a comparable pattern emerges. Nevertheless, our proposed method exhibits marginally elevated values for the unlearning class accuracy ($A_{D_f}$) in contrast to other baseline methods. This minor increase in $A_{D_f}$ is accompanied by a noteworthy trade-off, manifesting as a substantial improvement of 5\% to 20\% in retained class accuracy ($A_{D_r}$) when compared to the baseline methods. This indicates that, while our method may have a slight compromise in unlearning effectiveness on the specified class, it compensates for this by significantly enhancing the preservation of knowledge pertaining to the retained classes. The observed trade-off underscores the efficacy of our proposed method in achieving a more favorable balance between unlearning and retaining class-specific information when compared to existing baselines on the CIFAR-10, and CIFAR-100 datasets.
\begin{wraptable}{l}{0.42\textwidth}
    \centering
    \resizebox{0.42\textwidth}{!}{
    \begin{tabular}{c|ccccc}
    \toprule
    \textbf{Models} & \textbf{Datasets} & \textbf{|Unlearning Classes|} & \textbf{MIA accuracy} \\
    \midrule
    \multirow{4}{*}{Resnet-18} & MNIST & Class-2 & 0.24$\pm$0.01 \\
    & CIFAR-10 & Class-3 & 0.49$\pm$0.01 \\
    & CIFAR-100 & Class-1 & 0.48$\pm$0.03 \\
    & Food-101 & Class-8 & 0.49$\pm$0 \\
    \midrule
    \multirow{4}{*}{Resnet-34} & MNIST & Class-8 & 0.42$\pm$0.02 \\
    & Cifar-10 & Class-1 & 0.44$\pm$0 \\
    & Cifar-100 & Class-3 & 0.44$\pm$0.01 \\
    & Food-101 & Class-40 & 0.49$\pm$0 \\
    \midrule
    \multirow{2}{*}{Resnet-50} & Cifar-100 & Class-8 & 0.49$\pm$0 \\
    & Food-101 & Class-50 & 0.5$\pm$0 \\
    \midrule
    \multirow{2}{*}{All-CNN} & MNIST & Class-5 & 0.48$\pm$0.02 \\
    & Cifar-10 & Class-6 & 0.48$\pm$0.01 \\
\bottomrule
\end{tabular}
}
\caption{Membership Inference Attack(MIA) accuracy of the unlearned models}
\label{tab:MIA}
\end{wraptable}
Similarly, for the FOOD-101 dataset, our proposed method demonstrates a comparable accuracy on the unlearning class ($A_{D_f}$) when juxtaposed with other baseline methods. However, a noteworthy distinction arises in the accuracy of the retrained classes ($A_{D_r}$), where our method excels by exhibiting an improvement of 5\% to 10\% over the baseline approaches.

As previously elucidated, the Membership Inference Attack (MIA) accuracy stands as a pivotal criterion in evaluating unlearning methodologies. It is imperative for any unlearning mechanism to avoid escalating the MIA accuracy beyond the level achievable by random chance. The observations from Table-\ref{tab:MIA} underscore that our proposed unlearning method consistently maintains MIA accuracy below 0.5. 

\begin{wrapfigure}{r}{0.47\textwidth}
    \centering
    \includegraphics[width=0.47\textwidth]{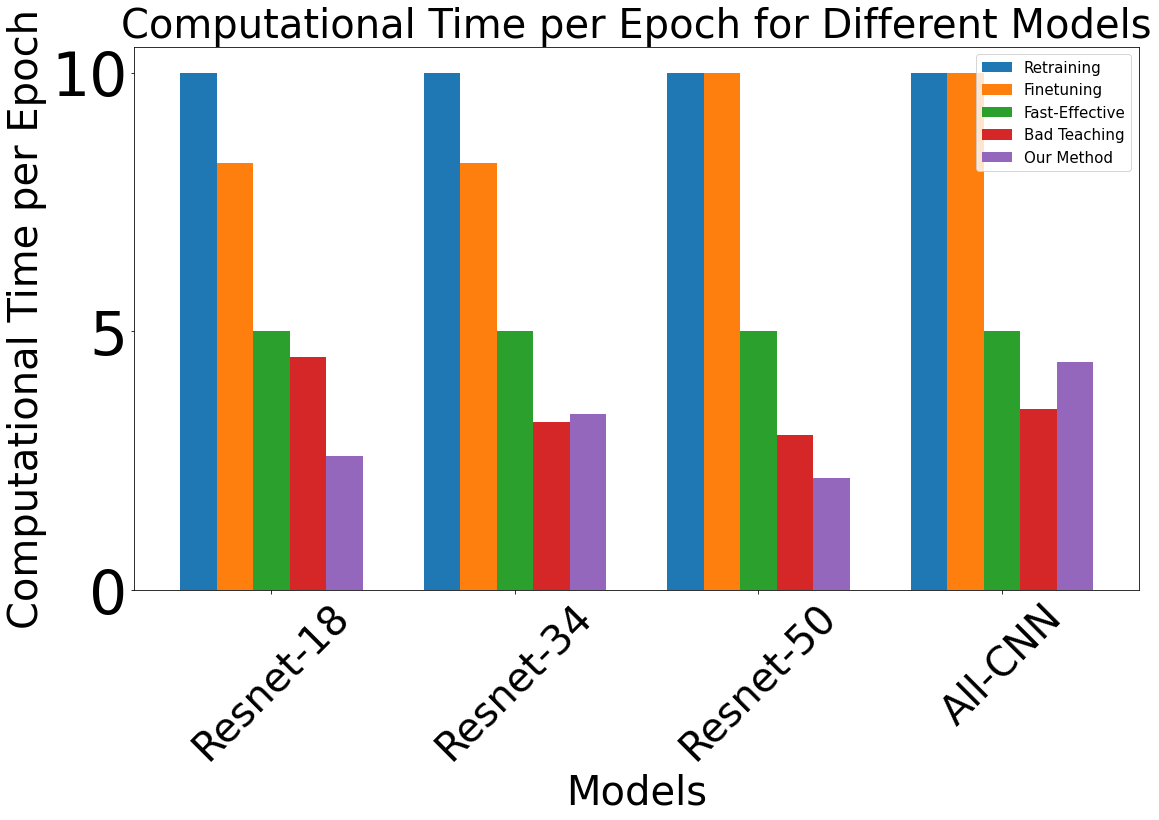}
    \caption{Unlearning Time comparison of our method with the Fast Yet Effective and Bad Teaching approach.}
    \label{fig:enter-label}
\end{wrapfigure}

In a broader context, MIA accuracy manifests variability contingent upon the specific model, dataset, and unlearning class under consideration. Resnet-18 exhibits a propensity for lower MIA accuracy when applied to the MNIST dataset. Conversely, Resnet-34 demonstrates relatively consistent performance across diverse datasets. Resnet-50 and All-CNN consistently register MIA accuracy below 0.5, indicative of the effectiveness of our devised unlearning algorithm. 

The essence of unlearning resides in its capacity to emulate the performance attained through retraining while also improving computational efficiency. As depicted in Fig.-\ref{fig:enter-label}, our proposed method yields superior results with fewer epochs. Notably, our approach is a single-step method, eliminating the need for a Repair step as seen in alternative approaches~\cite{tarun2023fast}. This dual accomplishment of improved performance and decreased computational burden positions our method as a practical and efficient solution for class unlearning tasks.


\subsection{Ablation Study}

\begin{table*}[!b]{}
\begin{minipage}{0.5\textwidth}
    \resizebox{\textwidth}{!}{
    \begin{tabular}{cccc|ccc}
    \toprule
    \textbf{Model} & \textbf{Dataset} & \textbf{Class} & \boldmath{($\alpha, \beta, \gamma$)} & \boldmath{$A_{D_f}$} & \boldmath{$A_{D_r}$} \\
    \midrule
    \textbf{Resnet-18} & MNIST & Class-2 & (631,0,0) & 0$\pm$0 & 41.45$\pm$6.22 \\
    \textbf{All-CNN} & CIFAR-10 & Class-3 &(2400,0,0) &  7.4$\pm$3.08 & 60.59$\pm$3.71 \\
    \textbf{Resnet-50} & CIFAR-100 & Class-3 & (91,0,0) & 0$\pm$0 & 15.83$\pm$15.35 \\
    \textbf{Resnet-50} & Food-101 & Class-30 & (721,0,0) & 0$\pm$0 & 15.09$\pm$10.19 \\
    \bottomrule
    \end{tabular}
    }
    \caption{Effect of the stability regularizer on $A_{D_f}$ and $A_{D_r}$: setting $\alpha$ to optimal and $\beta,\gamma$ to zero}
    \label{tab:effect-regularization}
\end{minipage}~
\begin{minipage}{0.5\textwidth}
    \resizebox{\textwidth}{!}{
    \begin{tabular}{ccc|cccc}
    \toprule
    \textbf{Model} & \textbf{Dataset} & \textbf{Unlearn-Class} & \boldmath{$\alpha$} & \boldmath{$A_{D_f}$} & \boldmath{$A_{D_r}$} \\
    \midrule
    \multirow{3}{*}{\textbf{Resnet-18}} & \multirow{3}{*}{\textbf{MNIST}} & \multirow{3}{*}{\textbf{Class-2}}  & 100 & 5.85$\pm$0.48 & 95.95$\pm$0.16 \\
    & & &200 & 0.1$\pm$0.1 & 96.74$\pm$0.16 \\
    & & & 400 & 0$\pm$0 & 97.24$\pm$0.3 \\
    & & & 800 & 0$\pm$0 & 97.16$\pm$0.68 \\
    \midrule
    \multirow{3}{*}{\textbf{All-CNN}} & \multirow{3}{*}{\textbf{CIFAR-10}} & \multirow{3}{*}{\textbf{Class-3}} & 500 & 6.53$\pm$1.85 & 83.64$\pm$0.31 \\
    & & & 1000 & 3.57$\pm$1.62 & 83.25$\pm$0.54 \\
    & & & 2000 & 0$\pm$0 & 81.23$\pm$0.48 \\
    & & & 4000 & 0.03$\pm$0.06 & 75.73$\pm$0.45 \\
    \midrule
    \multirow{3}{*}{\textbf{Resnet-50}} & \multirow{3}{*}{\textbf{CIFAR-100}} & \multirow{3}{*}{\textbf{Class-3}} & 40 & 1.33$\pm$0.58 & 72.35$\pm$1.87 \\
    & & & 80 & 0.33$\pm$0.58 & 72.72$\pm$0.57 \\
    & & & 160 & 0$\pm$0 & 71.84$\pm$0.61 \\
    & & & 320 & 0$\pm$0 & 71.52$\pm$0.24 \\
    \midrule
    \multirow{3}{*}{\textbf{Resnet-50}} & \multirow{3}{*}{\textbf{Food-101}} & \multirow{3}{*}{\textbf{Class-30}} & 40 & 1.87$\pm$1.15 & 70.27$\pm$0.64 \\
    & & & 80 & 0.4$\pm$0 & 65.52$\pm$0 \\
    & & & 160 & 0.8$\pm$1.39 & 69.64$\pm$0.48 \\
    & & & 320 & 0.93$\pm$1.62 & 67.77$\pm$1.21 \\
    \bottomrule
    \end{tabular}
}
\caption{Effect of $\alpha$ on $A_{D_f}$ and $A_{D_r}$: Setting $\beta,\gamma$ to optimal and varying $\alpha$}
\label{tab:effect-alpha}
\end{minipage}
\end{table*}

Table-\ref{tab:effect-regularization} presents valuable insights into the impact of a stability regularizer on different models across diverse datasets and classes. The absence of a stability regularizer is noteworthy, revealing that while the accuracy on the unlearned class ($A_{D_f}$) is exceptionally low (approaching 0), the accuracy of the model on the retained classes ($A_{D_r}$) experiences a significant decline. The findings emphasize that, without adequate stability regularization, models may struggle to maintain performance in the classes they were initially trained on. The experimental results indicate consistent trends across various settings, showcasing that the inclusion of a stability regularizer leads to a substantial improvement in the accuracy of retained classes. The observed gain in retain accuracy ranges from 20\% to 50\% under different experimental conditions.

Further to delve deeper into the influence of the parameter $\alpha$, we conducted additional experiments where we held the values of $\beta$ and $\gamma$ constant at their optimal settings while systematically varying the value of $\alpha$. The results, as presented in Table-\ref{tab:effect-alpha}, reveal intriguing insights. As the value of $\alpha$ is progressively increased, a discernible pattern emerges. Specifically, there is a consistent reduction in accuracy on the unlearned class ($A_{D_f}$) under the same number of training epochs. However, in contrast, the accuracy of the retained class ($A_{D_r}$) exhibits a declining trend.

\section{Conclusion}
We present a novel method tailored for unlearning specific classes within deep classification models. A key distinguishing feature of our approach is its capability to function effectively even with partial access only to the unlearning class data. Unlike existing methods that may require multiple steps or complete dataset access for unlearning, our method achieves its objective in a single step. We believe that this unique characteristic significantly enhances the practicality and efficiency of the unlearning process, particularly in scenarios where full dataset access is restricted or impractical. Noteworthy is the method's exceptional ability to surpass contemporaneous approaches consistently across a range of class unlearning challenges, underscoring its versatility and effectiveness. Our work represents a significant advancement in addressing fundamental challenges in machine learning related to privacy, safety, and model adaptability, particularly in the context of growing concerns about data privacy and model biases.

\bibliographystyle{splncs04}
\bibliography{samplepaper}

\end{document}


\title{Partially Blinded Unlearning: Appendix}

\maketitle
\section{Dataset and Model Details}
Here we have used 4 datasets as follows:
\begin{itemize}
    \item \textbf{MNIST:} The MNIST dataset consist of 28 $\times$ 28 grayscale representing handwritten digits from 0 to 9. The MNIST dataset contains 60,000 training images and 10,000 testing images, 
    \item  \textbf{CIFAR-10:} CIFAR10 consists of 60,000 32 $\times$ 32 color images distributed across 10 classes with 6000 images in each class.
    \item \textbf{CIFAR-100:}  CIFAR-100 is a benchmark dataset for object classification tasks, consisting of 60,000, 32 $\times$ 32 color images across 100 classes where 50,000 images are used for training and 10,000 images are used for testing. Each class contains 600 images, split into 500 for training and 100 for testing.
    \item \textbf{Food-101:} The Food101 consist of food images with $512\times512$ resolution that are allocated over 101 classes with each class containing 1000 images. A total of 75,750 images are used for training and 25,250 images are used for testing.
\end{itemize}

For Initial Training and Unlearning we have used 4 Models as follows:
\begin{itemize}
    \item \textbf{All-CNN:} All layers in the AllCNN architecture are convolutional layers. These layers consist of a set of learnable filters (also called kernels) that slide over the input image, performing convolution operations to extract features. Each filter detects different patterns in the input data. ReLU activation functions are applied after each convolutional layer to introduce non-linearity into the network. ReLU activation helps the network learn complex relationships in the data and speeds up the training process by allowing for faster convergence. Pooling layers are used to downsample the feature maps obtained from the convolutional layers, reducing the spatial dimensions of the data while retaining important features. Max pooling is a common choice, where the maximum value within a region of the feature map is retained, effectively reducing its size. nstead of using fully connected layers for classification, the AllCNN architecture employs global average pooling. This layer computes the average value of each feature map across its entire spatial dimensions. The resulting vector is then fed directly into the final softmax layer for classification. Global average pooling reduces the number of parameters in the network and helps prevent overfitting.
    \item \textbf{Resnet-18:} ResNet-18 is a convolutional neural network that is 18 layers including convolutional layers, batch normalization layers, ReLU activation functions, max-pooling layers, and fully connected layers. The network architecture can be divided into multiple stages, each containing several residual blocks with different numbers of filters. The first stage typically includes a single convolutional layer followed by several residual blocks, while subsequent stages repeat this pattern with gradually decreasing spatial dimensions and increasing number of filters.
    \item \textbf{Resnet-34:} ResNet34 consists of 34 layers, hence the name. These layers include convolutional layers, pooling layers, activation functions, and fully connected layers at the end for classification. ResNet34 primarily employs 3$\times$3 convolutional filters. Each residual block in ResNet34 typically contains two or more convolutional layers. These convolutional layers are followed by batch normalization and ReLU activation functions. ResNet34 uses max-pooling layers with a stride of 2 for downsampling feature maps spatially. These pooling layers help reduce the spatial dimensions of the feature maps while retaining the most important information.
    
    \item \textbf{Renset-50:} The ResNet50 architecture comprises four primary components: convolutional layers, identity blocks, convolutional blocks, and fully connected layers. Initially, convolutional layers extract features from input images. Subsequently, identity and convolutional blocks process and modify these features. Finally, fully connected layers perform the ultimate classification task. Within ResNet50, the convolutional layers consist of multiple convolutional operations, followed by batch normalization and ReLU activation. These layers extract diverse features like edges, textures, and shapes from input images. Following the convolutional layers, max-pooling layers are employed to downscale the feature maps while retaining essential features. The core elements of ResNet50 are the identity block and convolutional block. The former involves passing input through convolutional layers and adding it back to the output, facilitating the learning of residual functions to achieve desired outputs. The convolutional block resembles the identity block but incorporates a 1x1 convolutional layer to reduce filter numbers before the 3$\times$3 convolutional layer. The final phase of ResNet50 involves fully connected layers, which are pivotal for classification. These layers produce the ultimate classification through the application of a softmax activation function to the output of the final fully connected layer, generating class probabilities.
\end{itemize}

\section{Initial Training}
The below Table shows the experimental setting for initial training across different models and datasets.
For all experimental settings We have used Adam optimizer.

\begin{table}[htbp]
    \centering
    \caption{Experimental Settings for \textbf{Initial Training} of different models on different datasets}
    \label{tab:experimental_settings}
    \resizebox{\textwidth}{!}{
    \begin{tabular}{cc|cccc}
    \toprule
    \textbf{Dataset} & \textbf{Models} & \textbf{Epochs} & \textbf{Learning rate} & \textbf{Batch size} & \textbf{Other Hyper-parameters} \\
    \midrule
    \multirow{3}{*}{\textbf{MNIST}} & Resnet-18 & 10 & 0.0001 & 64 & None \\
    & All-CNN & 10 & 0.001 & 64 & None \\
    & Resnet-34 & 10 & 0.0001 & 48 & None \\
    \midrule
    \multirow{3}{*}{\textbf{Cifar-10}} & Resnet-18 & 10 & 0.001 & 64 & None \\
    & All-CNN & 10 & 0.001 & 64 & None \\
    & Resnet-34 & 10 & 0.001 & 64 & None \\
    \midrule
    \multirow{3}{*}{\textbf{Cifar-100}} & Resnet-18 & 10 & 0.0002 & 64 & betas=(0.9,0.999), eps=1e-8, Linear Scheduler \\
    & Rensnet-50 & 10 & 0.0002 & 64 & betas=(0.9,0.999), eps=1e-8, Linear Scheduler \\
    & Resnet-34 & 10 & 0.0002 & 64 & betas=(0.9,0.999), eps=1e-8, Linear Scheduler \\
    \midrule
    \multirow{3}{*}{\textbf{Food-101}} & Resnet-18 & 10 & 0.0001 & 64 & None \\
    & Resnet-50 & 10 & 0.0001 & 64 & None \\
    & Resnet-34 & 10 & 0.0003 & 64 & None \\
    \bottomrule
    \end{tabular}
    }
\end{table}

\section{Baselines for Unlearning}
Here we train four base-lines method that as mentioned in the paper. All the experimental details for these baselines are given for four datasets used. 

\begin{itemize}
    \item \textbf{Retraining:} We trained the model anew using data excluding the class designated for unlearning, employing 10 epochs, a batch size of 64, and a learning rate of 0.0001. The retrained model exhibited 0\% accuracy on the unlearned class, in contrast to the initial model's significant accuracy on the same class, while maintaining similar accuracy on the retained classes.
    \item \textbf{Fine-Tuning:} Rather than retraining from scratch, we initialized the model from the initial pre-trained checkpoint and fine-tuned it using the retained class datapoints. This process was executed for 10 epochs, utilizing a batch size of 64 and a learning rate of 0.0002.
    \item \textbf{Fast-Effective:} We conducted a comparative analysis between our model and the Fast Yet Effective approach for unlearning. In this method, one needs to maximize noise by employing a noise matrix to manipulate the weights of the initial model for unlearning through impair-repair steps, where weights are manipulated in the impair step and stabilized on the retained classes in the subsequent repair step. Our initial model was trained using this unlearning approach with a maximum learning rate of 0.001, weight decay fixed at $10^{-4}$, and a batch size of 64 for 10 epochs. It is noteworthy that this approach typically requires an average of 5 epochs to attain optimal performance of the unlearned model.
    \item \textbf{Bad Teaching:} Additionally, our method was compared with the bad teaching approach. In this method, a competent teacher, an incompetent teacher, and a student model are involved, where the student model attempts to unlearn classes specified by the user based on the information received from both teachers. The incompetent teacher's role is to mislead the student with incorrect information on the unlearning class data while the competent teacher is used to retain the information on the retained class data. All models were trained for 10 epochs with a fixed learning rate of 0.0001, and the hyper-parameter value of KL temperature was fixed at 1. This method typically requires approximately 3-4 epochs to achieve optimal performance.
\end{itemize}

\begin{table}[htbp]
    \centering
    \caption{Experimental Settings for Baseline Method on \textbf{MNIST} dataset}
    \label{tab:experimental_settings}
    \resizebox{\textwidth}{!}{
    \begin{tabular}{cc|cccc}
    \toprule
    \textbf{Methods} & \textbf{Models} &\textbf{ Epochs} & \textbf{Learning rate} & \textbf{Batch size} & \textbf{Other Hyper-parameters} \\
    \midrule
    \multirow{3}{*}{\textbf{Retraining}} & Resnet-18 & 10 & 0.001 & 64 & None \\
               & All-CNN & 10 & 0.001 & 64 & None \\
               & Resent-34 & 10 & 0.001 & 64 & None \\
    \midrule
    \multirow{3}{*}{\textbf{Finetuning}} & Resnet-18 & 10 & 0.001 & 64 & None \\
               & All-CNN & 10 & 0.001 & 64 & None \\
               & Resent-34 & 10 & 0.001 & 64 & None \\
    \midrule
    \multirow{3}{*}{\textbf{Fast-Effective}} & Resnet-18 & 5 & 0.1 & 64 & weight decay=1e-4, grad clip=0.1 \\
                   & All-CNN & 5 & 0.1 & 64 & weight decay=1e-4, grad clip=0.1 \\
                   & Resnet-34 & 5 & 0.001 & 64 & weight decay=1e-4, grad clip=0.1 \\
    \midrule
    \multirow{3}{*}{\textbf{Bad-Teaching}} & Resnet-18 & 3 & 0.001 & 256 & KL temperature=1 \\
                 & All-CNN & 3 & 0.001 & 64 & KL temperature=1 \\
                 & Resnet-34 & 3 & 0.001 & 64 & KL temperature=1 \\
    \bottomrule
    \end{tabular}
    }
\end{table}

\begin{table}[htbp]
    \centering
    \caption{Experimental Settings for Baseline Method on \textbf{Cifar-10} dataset}
    \label{tab:experimental_settings}
    \resizebox{\textwidth}{!}{
    \begin{tabular}{cc|cccc}
    \toprule
    \textbf{Methods} & \textbf{Models} &\textbf{ Epochs} & \textbf{Learning rate} & \textbf{Batch size} & \textbf{Other Hyper-parameters }\\
    \midrule
    \multirow{3}{*}{\textbf{Retraining}} & Resnet-18 & 10 & 0.001 & 64 & None \\
               & All-CNN & 10 & 0.001 & 64 & None \\
               & Resnet-34 & 10 & 0.001 & 64 & None \\
    \midrule
    \multirow{3}{*}{\textbf{Finetuning}} & Resnet-18 & 10 & 0.001 & 64 & None \\
               & All-CNN & 10 & 0.001 & 64 & None \\
               & Resnet-34 & 10 & 0.001 & 64 & None \\
    \midrule
    \multirow{3}{*}{\textbf{Fast-Effective}} & Resnet-18 & 5 & 0.001 & 64 & weight decay=1e-4, grad clip=0.1 \\
                   & All-CNN & 5 & 0.001 & 64 & weight decay=1e-4, grad clip=0.1 \\
                   & Resnet-34 & 5 & 0.001 & 64 & weight decay=1e-4, grad clip=0.1 \\
    \midrule
    \multirow{3}{*}{\textbf{Bad-Teaching}} & Resnet-18 & 10 & 0.0001 & 256 & KL=1 \\
                 & All-CNN & 3 & 0.0001 & 64 & KL=1 \\
                 & Resnet-34 & 2 & 0.0001 & 64 & KL=1 \\
    \bottomrule
    \end{tabular}
    }
\end{table}

\begin{table}[htbp]
    \centering
    \caption{Experimental Settings for Baseline Method on \textbf{CIFAR-100} dataset}
    \label{tab:experimental_settings}
    \resizebox{\textwidth}{!}{
    \begin{tabular}{cc|cccc}
    \toprule
    \textbf{Methods} &\textbf{ Models} & \textbf{Epochs} & \textbf{Learning rate} & \textbf{Batch size} & \textbf{Other Hyper-parameters} \\
    \midrule
    \multirow{3}{*}{\textbf{Retraining}} & Resnet-18 & 10 & 0.0002 & 64 & betas=(0.9,0.999), eps=10e-8, linear lr schedulder \\
               & Renset-50 & 10 & 0.0002 & 64 & betas=(0.9,0.999), eps=10e-8, linear lr schedulder \\
               & Resnet-34 & 10 & 0.0002 & 64 & betas=(0.9,0.999), eps=10e-8, linear lr schedulder \\
    \midrule
    \multirow{3}{*}{\textbf{Finetuning}} & Resnet-18 & 10 & 0.0002 & 64 & betas=(0.9,0.999), eps=10e-8, linear lr schedulder \\
               & Resnet-50 & 10 & 0.0002 & 64 & betas=(0.9,0.999), eps=10e-8, linear lr schedulder \\
               & Resnet-34 & 10 & 0.0002 & 64 & betas=(0.9,0.999), eps=10e-8, linear lr schedulder \\
    \midrule
    \multirow{3}{*}{Fast-Effective} & Resnet-18 & 5 & 0.001 & 64 & weight decay=1e-4, grad clip=0.1 \\
                   & Resnet-50 & 5 & 0.001 & 64 & weight decay=1e-4, grad clip=0.1 \\
                   & Resnet-34 & 5 & 0.001 & 64 & weight decay=1e-4, grad clip=0.1 \\
    \midrule
    \multirow{3}{*}{\textbf{Bad-Teaching}} & Resnet-18 & 3 & 0.001 & 256 & Kl=1, betas=(0.9,0.999), eps=10e-8 \\
                 & Resnet-50 & 3 & 0.0001 & 256 & Kl=1, betas=(0.9,0.999), eps=10e-8 \\
                 & Resnet-34 & 3 & 0.0001 & 256 & Kl=1, betas=(0.9,0.999), eps=10e-8 \\
    \bottomrule
    \end{tabular}
    }
\end{table}

\begin{table}[htbp]
    \centering
    \caption{Experimental Settings for Baseline Method on \textbf{Food-101} dataset}
    \label{tab:experimental_settings_L}
    \resizebox{\textwidth}{!}{
    \begin{tabular}{cc|cccc}
    \toprule
    \textbf{Methods} & \textbf{Models} & \textbf{Epochs} & \textbf{Learning rate} & \textbf{Batch size} & \textbf{Other Hyper-parameters} \\
    \midrule
    \multirow{3}{*}{\textbf{Retraining}} & Resnet-18 & 10 & 0.0002 & 64 & betas=(0.9,0.999), eps=10e-8, linear learning rate \\
               & Resnet-50 & 10 & 0.0002 & 64 & betas=(0.9,0.999), eps=10e-8, linear learning rate \\
               & Resnet-34 & 10 & 0.0002 & 64 & linear learning rate scheduler \\
    \midrule
    \multirow{3}{*}{\textbf{Finetuning}} & Resnet-18 & 3 & 0.0002 & 64 & betas=(0.9,0.999), eps=10e-8 \\
               & Resnet-50 & 3 & 0.0001 & 64 & linear learning rate scheduler \\
               & Resnet-34 & 3 & 0.0001 & 64 & linear learning rate scheduler \\
    \midrule
    \multirow{3}{*}{Fast-Effective} & Resnet-18 & 5 & 0.001 & 64 & weight decay=1e-4, grad clip=0.1 \\
                   & Resnet-50 & 5 & 0.001 & 64 & weight decay=1e-4, grad clip=0.1 \\
                   & Resnet-34 & 5 & 0.001 & 64 & weight decay=1e-4, grad clip=0.1 \\
    \midrule
    \multirow{3}{*}{Bad-Teaching} & Resnet-18 & 2 & 0.001 & 64 & KL=1 \\
                 & Resnet-50 & 2 & 0.001 & 64 & KL=1 \\
                 & Resnet-34 & 2 & 0.001 & 64 & KL=1 \\
    \bottomrule
    \end{tabular}
    }
\end{table}

\newpage

\section{PBU(Our Method)}
In this method, we minimize the following loss function $\mathcal{L}(\theta, \theta^*, \mathcal{S}_n)$, where 
\begin{equation}
    \mathcal{L}(\theta, \theta^*, \mathcal{S}_n) = \alpha \log P(\mathcal{S}_n|\theta) + \beta (\theta - \theta^*)^T I_{\theta^*}(\mathcal{S}_n)(\theta - \theta^*) + \gamma ||\theta - \theta^*||^2 
    \label{eq-10}
\end{equation}

Here $\alpha, \beta, \gamma$ are hyper-parameters and $I_{\theta^*}(\mathcal{S}_n)$ is the Fisher Information Matrix of the initial parameters corresponding to the unlearning data class. The overall methodology, entails perturbing the initial parameter $\theta^*$ to the unlearned parameter $\theta^u$ using the loss function described in Equation-\ref{eq-10}. This loss function comprises two components: the first component involves perturbing the parameters by minimizing the log-likelihood associated with the unlearning class data, while the second component involves perturbing parameters by incorporating stability regularization in the parameter space. This stability regularization encompasses the Mahalanobis distance between the initial parameter and the unlearned parameter with respect to the Fisher Information matrix corresponding to the negative class, along with the $l_2$ distance between the initial parameter and the unlearned parameter.

\begin{table}[!]
    \centering
    \caption{Experimental Settings for PBU Method on \textbf{MNIST} dataset}
    \label{tab:experimental_settings}
    \resizebox{\textwidth}{!}{
    \begin{tabular}{cc|ccccc}
    \toprule
    Models & Classes & Epochs & Learning Rate & Batch Size & $(\alpha,\beta,\gamma)$ & Other Hyperparameters\\
    \midrule
    \multirow{3}{*}{\textbf{Resnet-18}} & Class-2 & 3 & 0.0001 & 64 & (631,100,4001) & None \\
    & Class-6 & 3 & 0.0001 & 64 & (721,100,6001) & None \\
    & Class-8 & 2 & 0.0001 & 64 & (671,100,4001) & None\\
    \midrule
    \multirow{3}{*}{\textbf{All-CNN}} & Class-2 & 3 & 0.001 & 64 & (811,100,4001) & None \\
    & Class-6 & 1 & 0.001 & 64 & (631,100,4001) & None\\
    & Class-8 & 1 & 0.001 & 64 & (721,100,4001)  & None\\
    \midrule
    \multirow{3}{*}{\textbf{Resent-34}} & Class-2 & 1 & 0.0001 & 64 & (721,100,4002) & None \\
    & Class-6 & 1 & 0.0002 & 64 & (621,100,4001) & None\\
    & Class-8 & 1 & 0.0001 & 64 & (621,100,4002) & None \\
    \bottomrule
    \end{tabular}
    }
\end{table}

\begin{table}[!]
    \centering
    \caption{Experimental Settings for PBU Method on \textbf{CIFAR10} dataset}
    \label{tab:experimental_settings}
    \resizebox{\textwidth}{!}{
    \begin{tabular}{cc|ccccc}
    \toprule
    Models & Classes & Epochs & Learning Rate & Batch Size & $(\alpha,\beta,\gamma)$ & Other Hyperparameters\\
    \midrule
    \multirow{3}{*}{\textbf{Resnet-18}} & Class-1 & 3 & 0.002 & 64 & (1501,100,15000) & betas=(0.9,0.999), eps=10e-8\\
    & Class-3 & 3 & 0.0001 & 64 & (91,100,20000) & None \\
    & Class-5 & 5 & 0.0001 & 64 & (361,100,20000) & None\\
    \midrule
    \multirow{3}{*}{\textbf{All-CNN}} & Class-1 & 4 & 0.0001 & 64 & (2400,88,14000) & None \\
    & Class-3 & 10 & 0.0001 & 64 & (2400,88,14000) & None\\
    & Class-5 & 8 & 0.0001 & 64 & (2400,88,14000)  & None\\
    \midrule
    \multirow{3}{*}{\textbf{Resent-34}} & Class-1 & 3 & 0.0001 & 64 & (721,100,4001) & None \\
    & Class-3 & 3 & 0.0001 & 64 & (721,100,4001)  & None\\
    & Class-5 & 2 & 0.0001 & 64 & (721,100,4001)  & None \\
    \bottomrule
    \end{tabular}
    }
\end{table}

\begin{table}[!]
    \centering
    \caption{Experimental Settings for PBU Method on \textbf{CIFAR100} dataset}
    \label{tab:experimental_settings}
    \resizebox{\textwidth}{!}{
    \begin{tabular}{cc|ccccc}
    \toprule
    Models & Classes & Epochs & Learning Rate & Batch Size & $(\alpha,\beta,\gamma)$ & Other Hyperparameters\\
    \midrule
    \multirow{3}{*}{\textbf{Resnet-18}} & Class-1 & 4 & 0.0001 & 64 & (91,100,4001) & None\\
    & Class-3 & 7 & 0.0001 & 64 & (91,90,4001) & None \\
    & Class-5 & 7 & 0.0001 & 64 & (91,100,4000) & None\\
    \midrule
    \multirow{3}{*}{\textbf{Resnet-50}} & Class-1 & 2 & 0.0001 & 64 & (271,100,4001) & betas=(0.9,0.999), eps=10e-8, linear lr schedulder \\
    & Class-3 & 2 & 0.0001 & 64 &  (91,90,5001) & betas=(0.9,0.999), eps=10e-8, linear lr schedulder\\
    & Class-5 & 3 & 0.0001 & 64 & (91,90,5001)  & betas=(0.9,0.999), eps=10e-8, linear lr schedulder\\
    \midrule
    \multirow{3}{*}{\textbf{Resent-34}} & Class-1 & 4 & 0.0001 & 64 & (1521,80,6001) & None \\
    & Class-3 & 3 & 0.0001 & 64 & (1001,80,6001)  & None\\
    & Class-5 & 4 & 0.0001 & 64 & (1521,80,6001)  & None \\
    \bottomrule
    \end{tabular}
    }
\end{table}

\begin{table}[htbp]
    \centering
    \caption{Experimental Settings for PBU Method on \textbf{Food101} dataset}
    \label{tab:experimental_settings}
    \resizebox{\textwidth}{!}{%
    \begin{tabular}{cc|ccccc}
    \toprule
    Models & Classes & Epochs & Learning Rate & Batch Size & $(\alpha,\beta,\gamma)$ & Other Hyperparameters\\
    \midrule
    \multirow{3}{*}{\textbf{Resnet-18}} & Class-1 & 1 & 0.0002 & 64 & (4.8,135.12,24688) & None\\
    & Class-3 & 1 & 0.0002 & 64 & (721,20015,5001) & None \\
    & Class-5 & 2 & 0.0002 & 64 & (721,20015,5001) & None\\
    \midrule
    \multirow{3}{*}{\textbf{Resnet-50}} & Class-1 & 2 & 0.0002 & 64 & (721,20015,5000) & None \\
    & Class-3 & 1 & 0.0002 & 64 &  (721,20015,5000) & None \\
    & Class-5 & 1 & 0.0002 & 64 & (721,20015,5000) & None \\
    \midrule
    \multirow{3}{*}{\textbf{Resent-34}} & Class-1 & 1 & 0.0003 & 64 & (721,20015,5000) & None \\
    & Class-3 & 1 & 0.0003 & 64 & (721,20015,5000) & None\\
    & Class-5 & 1 & 0.0003 & 64 & (721,20015,5000)  & None \\
    \bottomrule
    \end{tabular}%
    }
\end{table}

\newpage
\section{Unlearning Time}

\begin{table}[!]
    \centering
    \label{tab:experimental_settings}
    \resizebox{\textwidth}{!}{%
    \begin{tabular}{cc|cccc|c}
    \toprule
    \textbf{Models} & \textbf{Datasets} & \textbf{Retraining} & \textbf{Fine Tuning} & \textbf{Fast-Effective} & \textbf{Bad Teaching} & \textbf{Our Method} \\
    \midrule
    \multirow{4}{*}{\textbf{Resnet-18}} & MNIST & 10 & 10 & 5 & 3 & 2.11 \\
    & CIFAR-10 & 10 & 10 & 5 & 10 & 1.88 \\
    & CIFAR-100 & 10 & 10 & 5 & 1 & 5.11 \\
    & Food-101 & 10 & 3 & 5 & 4 & 1.3 \\
    \midrule
    \multirow{4}{*}{\textbf{Resnet-34}} & MNIST & 10 & 10 & 5 & 3 & 3 \\
    & CIFAR-10 & 10 & 10 & 5 & 3 & 3.7 \\
    & CIFAR-100 & 10 & 10 & 5 & 2 & 3.88 \\
    & Food-101 & 10 & 3 & 5 & 5 & 3 \\
    \midrule
    \multirow{2}{*}{\textbf{Resnet-50}} & CIFAR-100 & 10 & 10 & 5 & 1 & 3 \\
    & Food-101 & 10 & 10 & 5 & 5 & 1.33 \\
    \midrule
    \multirow{2}{*}{\textbf{All-CNN}} & MNIST & 10 & 10 & 5 & 4 & 3 \\
    & CIFAR-10 & 10 & 10 & 5 & 3 & 5.8 \\
    \bottomrule
    \end{tabular}%
    }
\caption{Average Unlearning Time for Different Classes across Different Models and Different Datasets}
\end{table}